%% file: main-storylines-2016.tex
\documentclass[11pt]{article}
\usepackage{emnlp2016}
\usepackage{times}
\usepackage{url}
\usepackage{tikz}
\usepackage{amsmath,amsfonts,amssymb}
\usepackage{booktabs}
\usepackage{multirow}

\makeatletter
\newcommand{\@BIBLABEL}{\@emptybiblabel}
\newcommand{\@emptybiblabel}[1]{}
\makeatother

\emnlpfinalcopy

\usepackage[draft,hidelinks]{hyperref}

\usepackage{bm}
\renewcommand{\vec}[1]{\bm{#1}}
\newcommand{\vw}[0]{\vec{w}}
\newcommand{\vc}[0]{\vec{c}}
\newcommand{\vzc}[0]{\vec{z}^{(\vc)}}
\newcommand{\zci}[0]{z_i^{(\vc)}}
\newcommand{\zc}[0]{z^{(\vc)}}
\newcommand{\citestr}{\bibliography{cite-strings,cites,cite-definitions}}

\title{Nonparametric Bayesian Storyline Detection from Microtexts}

\author{Vinodh Krishnan \\
  Georgia Institute of Technology \\
  Atlanta, GA 30308\\
  {\tt krishnan.vinodh@gmail.com} \\\And
  Jacob Eisenstein \\
  Georgia Institute of Technology \\
  Atlanta, GA 30308\\
  {\tt jacobe@gmail.com} \\}

\date{}

\begin{document}
\maketitle

\begin{abstract}
\input{abstract}
\end{abstract}

\input{intro}
\input{model}
\input{inference}
\input{eval-story}

\input{related}

\input{conclusion}
\input{ack}

\bibliographystyle{emnlp2016}
\citestr

\end{document}

%% file: abstract.tex
News events and social media are composed of evolving storylines, which capture public attention for a limited period of time. Identifying storylines requires integrating temporal and linguistic information, and prior work takes a largely heuristic approach. We present a novel online non-parametric Bayesian framework for storyline detection, using the distance-dependent Chinese Restaurant Process (dd-CRP). To ensure efficient linear-time inference, we employ a fixed-lag Gibbs sampling procedure, which is novel for the dd-CRP. We evaluate on the TREC Twitter Timeline Generation (TTG), obtaining encouraging results: despite using a weak baseline retrieval model, the dd-CRP story clustering method is competitive with the best entries in the 2014 TTG task.

%% file: intro.tex
\section{Introduction}
A long-standing goal for information retrieval and extraction is to identify and group textual references to ongoing events in the world~\cite{allan2002topic}. Success on this task would have applications in personalized news portals~\cite{gabrilovich2004newsjunkie}, intelligence analysis, disaster relief~\cite{vieweg2010microblogging}, and in understanding the properties of the news cycle~\cite{leskovec2009meme}. This task attains a new importance in the era of social media, where citizen journalists can document events as they unfold~\cite{lotan2011arab}, but where repetition and untrustworthy information can make the reader's task especially challenging~\cite{becker2011beyond,marcus2011twitinfo,petrovic2010streaming}.

A major technical challenge is in fusing information from two heterogeneous data sources: textual content and time. Two different documents about a single event might use very different vocabulary, particularly in sparse social media data such as microblogs; conversely, two different sporting events might be described in nearly identical language, with differences only in the numerical outcome. Temporal information is therefore critical: in the first case, to find the commonalities across disparate writing styles, and in the second case, to identify the differences. A further challenge is that unlike in standard document clustering tasks, the number of events in a data stream is typically unknown in advance. Finally, there is a high premium on scalability, since online text is produced at a high rate.

Due to these challenges, existing approaches for combining these modalities have been somewhat heuristic, relying on tunable parameters to control the tradeoff between textual and temporal similarity. In contrast, the Bayesian setting provides elegant formalisms for reasoning about latent structures (e.g., events) and their stochastically-generated realizations across text and time. In this paper, we describe one such model, based on the distance-dependent Chinese Restaurant Process (dd-CRP; Blei and Frazier, 2011)\nocite{blei2011distance}. This model is distinguished by the neat separation that it draws between textual content, which is treated as a stochastic emission from an unknown Multinomial distribution, and time, which is modeled as a prior on graphs over documents, through an arbitrary \emph{distance function}. However, straightforward implementations of the dd-CRP are insufficiently scalable, and so the model has been relatively underutilized in the NLP literature~\cite{titov2011bayesian,kim2011accounting,sirts2014pos}. We describe improvements to Bayesian inference that make the application of this model feasible, and present encouraging empirical results on the Tweet Timeline Generation task from TREC 2014~\cite{lin2014overview}.


%% file: model.tex
\section{Model}
The basic task that we address is to group short text documents into an unknown number of storylines, based on their textual content and their temporal signature. The textual content may be extremely sparse --- the typical Tweet is on the order of ten words long --- so leveraging temporal information is crucial. Moreover, the temporal signal is multiscale: in the 24-hour news cycle, some storylines last for less than an hour, while others, like the disappearance of the Malaysian Airlines 370 plane in 2014, continue for weeks or months. In some cases, the temporal distribution of references to a storyline will be unimodal and well-described by a parametric model~\cite{marcus2011twitinfo}; in other cases, it may be irregular, with bursts of activity followed by periods of silence~\cite{he2007analyzing}. Finally, it will be crucial to produce an implementation that scales to large corpora.

The distance-dependent Chinese Restaurant Process (dd-CRP) meets many of these criteria~\cite{blei2011distance}. In this model, the key idea is that each instance (document) $i$  ``follows'' another instance $c_i$ (where it is possible that $c_i = i$), inducing a graph. We can compute a partitioning over instances by considering the connected components in the undirected version of the follower graph; these partitions correspond to ``tables'' in the conventional ``Chinese Restaurant'' analogy~\cite{aldous1985exchangeability}, or to clusters. The advantage of this approach is that it is fundamentally non-parametric, and it introduces a clean separation between the textual data and the covariates: the text is generated by a distribution associated with the partition, while the covariates are associated with the following links, which are conditioned on a distance function. 

The distribution over follower links for document $i$ has the following form, 
\begin{equation}
\Pr(c_i = j) \propto
\begin{cases} 
  f(d_{i,j}), & i \neq j\\
  \alpha, & i = j,\\
\end{cases}
\label{eq:prior}
\end{equation}
where $d_{i,j}$ is the distance between units $i$ and $j$, and $\alpha > 0$ is a parameter of the model. Large values of $\alpha$ induce more self-links and therefore more fine-grained partitionings. Since we are concerned with temporal covariates, we define the distance function as follows: 
\begin{equation}
f(d_{i,j}) = e^{\frac{-|t_i - t_j|}{a}}.
\end{equation}
Thus, the likelihood of document $i$ following document $j$ decreases exponentially as the time gap $|t_i - t_j|$ increases.

The text of each document $i$ is represented by a vector of word counts $\vw_i$. The likelihood distribution is multinomial, conditioned on a parameter $\theta$ associated with the partition to which document $i$ belongs. By placing a Dirichlet prior on $\theta$, we can analytically integrate it out. Writing $\vzc$ for the cluster membership induced by the follower graph $\vc$, we have:
\begin{small}
\begin{align}
P(\vw \mid \vc; \eta) = & \prod_k P(\{ \vw_i : \zci = k \}; \eta)\\
 = &\prod_k \int_\theta P(\{ \vw_i : \zci = k \} \mid \theta) P(\theta ; \eta) d\theta
\label{eq:likelihood}
\end{align}
\end{small}

Given a multinomial likelihood $P(\vw \mid \theta)$ and a (symmetric) Dirichlet prior $P(\theta \mid \eta)$, this integral has a closed-form solution as the Dirichlet-Multinomial distribution (also known as the multivariate Polya distribution). The joint probability is therefore equal to the product of \autoref{eq:prior} and \autoref{eq:likelihood}, 
\begin{align}
P(\vw, \vc) = \prod_i P(c_i ; \alpha, a) \prod_k P(\{\vw_i : \zci = k\}; \eta).
\label{eq:joint}
\end{align}
The model has three hyperparameters: $\alpha$, which controls the likelihood of self-linking, and therefore affects the number of clusters; $a$, which controls the time scale of the distance function, and therefore affects the importance of the temporal dimension to the resulting clusters; and $\eta$, which controls the precision of the Dirichlet prior, and therefore the importance of rare words in the textual likelihood function. Estimation of these hyperparameters is described in \autoref{sec:hyper}.

%% file: inference.tex
\section{Inference}
\label{sec:inference}
The key sampling equation for the dd-CRP is the posterior likelihood,
\begin{align*}
\Pr(c_i = j \mid \vec{c}_{-i}, \vw) \propto & \Pr(c_i = j) P(\vw \mid \vec{c}).
\end{align*}
The prior is defined in Equation~\ref{eq:prior}. Let $\ell$ represent the likelihood under the partitioning induced when the link $c_i$ is cut. Now, the likelihood term has two cases: in the first case, $j$ is already in the same connected component as $i$ (even after cutting the link $c_i$), so no components are merged by setting $c_i = j$. In this case, the likelihood $P(\vw \mid c_i = j)$ is exactly equal to $\ell$. In the second case, setting $c_i = j$ causes two clusters to be merged. This gives the likelihood,
\begin{align*}
P&(\vw \mid c_i = j, \vec{c}_{-i}) \\
& \propto \frac{P(\{\vw_k : \zc_k = \zc_j \vee \zc_k = \zc_i\})}
{P(\{\vw_k : \zc_k = \zc_i\}) 
P(\{\vw_k : \zc_k = \zc_j\}) },
\label{eq:likelihood}
\end{align*}
where the constant of proportionality is exactly equal to $\ell$. Each of the terms in the likelihood ratio is a Dirichlet Compound Multinomial likelihood. This likelihood function is itself a ratio of gamma functions; by eliminating constant terms and exploiting the identity $\Gamma(x+1) = x \Gamma(x)$, we can reduce the number of Gamma function evaluations required to compute this ratio to the number of words which appear in \emph{both} clusters $\zc_i$ and $\zc_j$. Words that occur in neither cluster can safely be ignored, and the gamma functions for words which occur in exactly one of the two clusters cancel in the numerator and denominator of the ratio. Note also that we only need compute the likelihood for $c_i$ with respect to each cluster, not for every possible follower link. 

\subsection{Online inference}
\label{sec:online}
While we make every effort to accelerate the computation of individual Gibbs samples, the complexity of the basic algorithm is superlinear in the number of instances. This is due to the fact that each sample requires computing the probability of instance $i$ joining every possible cluster, while the number of clusters itself grows with the number of instances (this growth is logarithmic in the Chinese Restaurant Process). Scalability to the streaming setting therefore requires more aggressive optimizations.

To get back to linear time complexity, we employ a fixed-lag sampling procedure~\cite{doucet2000sequential}. After receiving instance $i$, we perform Gibbs sampling only within the fixed window $[t_i - \tau, t_i]$, leaving $c_j$ fixed if $t_j < t_i - \tau$. This approximate sampling procedure implicitly changes the underlying model, because there is no possibility of linking $i$ to a later message $j$ if the time gap $t_j - t_i > \tau$. 

Since we are only interested in obtaining a single storyline clustering --- rather than a full Bayesian distribution over clusterings --- we perform annealing for samples towards the end of the sampling window. Specifically, we set the temperature to $\gamma = 2.0$  and exponentiate the sampling likelihood by the inverse temperature~\cite{geman1984stochastic}. This has the effect of interpolating between probabilistically-correct Gibbs sampling and a hard coordinate-ascent procedure.

\subsection{Hyperparameter estimation}
\label{sec:hyper}
The model has three parameters to estimate: 
\begin{itemize}
\setlength\parskip{0pt}
\setlength\parsep{0pt}
\setlength\itemsep{0pt}
\item $\alpha$, the concentration parameter of the dd-CRP
\item $a$, the offset of the distance function
\item $\eta$, the scale of the symmetric Dirichlet prior.
\end{itemize}
We interleave maximization-based updates to these parameters with sampling, in a procedure inspired by Monte Carlo Expectation Maximization~\cite{wei1990monte}. Specifically, we compute gradients on the likelihood $P(\vc)$ with respect to $\alpha$ and $a$, and take gradient steps after every fixed number of samples. For the symmetric Dirichlet parameter $\eta$, we employ the heuristic from~\newcite{minka2000estimating} by setting the parameter to $\eta = \frac{(K-1)/2}{\sum_k\log p_k}$, where $K$ is the number of words that appear exactly once, and $p_k$ is the probability of choosing the $k^{th}$ word from the vocabulary under the unigram distribution for the entire corpus.

%% file: eval-story.tex
\section{TREC Evaluation}
To test the efficacy of this approach, we evaluate on the Twitter Timeline Generation (TTG) task in the Microblog track of TREC 2014. It involves taking tweets based on a query $Q$ at time $T$ and returning a summary that captures relevant information. We perform the task on 55 queries with different timestamps and compare our results with 13 groups that submitted 50 runs for this task in 2014. 

We consider the following systems:
\begin{description}
\setlength\parskip{0pt}
\setlength\parsep{0pt}
\item[Baseline] We replace the distance-dependent prior with a standard Dirichlet prior. The number of clusters is heuristically set to 20. Annealed Gibbs sampling is employed for inference.
\item[Offline inference] The dd-CRP model with offline inference procedure (described in \autoref{sec:inference}).
\item[Online inference] The dd-CRP model with online inference procedure (described in \autoref{sec:online}).
\end{description}

For the online inference implementation, we set the size of window and number of iterations to five days and 500 respectively. For the baseline, the parameter of the Dirichlet prior was set to a vector  of $0.5$ for each cluster. These values were chosen through 10-fold cross validation.

To measure the quality of the clusterings obtained by these models, we compare the average weighted and unweighted F-measures for 55 TREC topics, using the evaluation scripts from the TREC TTG task. Overall results are shown in \autoref{tab:sim}. The \textsc{online model} has the best weighted F1 score, outperforming the offline version of the same model, even though its inference procedure is an approximation to the \textsc{offline model}. It may be that its approximate inference procedure discourages long-range linkages, thus placing a greater emphasis on the temporal dimension. Both models were trained over 500 iterations, and the \textsc{online model} was 30\% faster to train than the offline model. 

Compared to the other 2014 TREC TTG systems, our dd-CRP models are competitive. Both models outperform all but one of the fourteen submissions on the unweighted $F_1$ metric, and would have placed fourth on the weighted $F^{w}_1$ metric. Note that the TREC evaluation scores both clustering quality and retrieval. We use only the baseline retrieval model, which achieved a mean average precision of 0.31. The competing systems shown in \autoref{tab:sim} all use retrieval models that are far superior: the retrieval model for top-ranked PKUICST team (line 4) achieved a mean average precision (MAP) of 0.59~\cite{lv2014pkuicst}, and the QCRI~\cite{magdy2014qcri} and and hltcoe~\cite{xu2014hltcoe} teams (lines 5 and 6) used retrieval models with MAP scores of at least 0.5. Bayesian dd-CRP storyline clustering was competitive with these timeline generation systems despite employing a far worse retrieval model, so improving the retrieval model to achieve parity with these alternative systems seems the most straightforward path towards better overall performance.

\begin{table*}
\centering
\begin{tabular}{l l l l l l}
    \toprule
    Model & Rec. & Rec.$^{w}$ & Prec. & $F_1$ & $F_1^{w}$ \\ \midrule
\textit{dd-CRP clustering models} \\
1. \textsc{baseline} & 0.14 & 0.27 & 0.33 & 0.20 & 0.30 \\
2. \textsc{offline} & 0.32 & 0.47 & 0.27 & 0.29 & 0.34 \\
3. \textsc{online} & 0.34 & 0.55 & 0.26 & 0.29 & 0.35 \\[8pt]
\textit{Top systems from Trec-2014 TTG}\\
4. \texttt{TTGPKUICST2}~\cite{lv2014pkuicst} & 0.37 & 0.58 & 0.46 & 0.35 & 0.46 \\
5. \texttt{EM50}~\cite{magdy2014qcri} & 0.29 & 0.48 & 0.42 & 0.25 & 0.38 \\
6. \texttt{hltcoeTTG1}~\cite{xu2014hltcoe} & 0.40 & 0.59 & 0.34 & 0.28 & 0.37 \\
    \bottomrule
\end{tabular}
\caption{Performance of Models in the TREC 2014 TTG Task. Weighted recall and $F_1$ are indicated as Rec.$^{w}$ and $F_1^w$.}
\label{tab:sim}
\end{table*}

%% file: related.tex
\section{Related work}
Topic tracking and first-story detection are very well-studied tasks; space does not permit a complete analysis of the related work, but see~\cite{allan2002topic} for a summary of ``first generation'' research. More recent non-Bayesian approaches have focused on string overlap~\cite{suen2013nifty}, submodular optimization~\cite{shahaf2012trains}, and locality-sensitive hashing~\cite{petrovic2010streaming}. 
In Bayesian storyline analysis, the seminal models are Topics-Over-Time~\cite{wang2006topics}, which associates a parametric distribution over time with each topic~\cite{ihler2006adaptive}, and the Dynamic Topic Model~\cite{blei2006dynamic}, which models topic evolution as a linear dynamical system~\cite{nallapati2007multiscale}. Later work by \newcite{diao2012finding} offers a model for identifying ``bursty'' topics, with inference requiring dynamic programming. All these approaches require the number of topics to be identified in advance. \newcite{kim2011accounting} apply a distance-dependent Chinese Restaurant \emph{Franchise} for temporal topic modeling; they evaluate using predictive likelihood rather than comparing against ground truth, and do not consider online inference.

The Infinite Topic-Cluster model~\cite{ahmed2011unified} is non-parametric over the number of storylines, through the use of the recurrent Chinese Restaurant Process (rCRP). The model is substantially more complex than our approach. Unlike the dd-CRP, the rCRP is Markovian in nature, so that the topic distribution at each point in time is conditioned on the previous epoch (or, at best, the previous $K$ epochs, with complexity of inference increasing with $K$). This Markovian assumption creates probabilistic dependencies between the topic assignment for a given document and the documents that follow in subsequent epochs, necessitating an inference procedure that combines sequential Monte Carlo and Metropolis Hastings, and a custom data structure; this inference procedure was complex enough to warrant a companion paper~\cite{ahmed2011online}. The rCRP is also employed by Diao and Jiang (2013, 2014)\nocite{diao2013unified,diao2014recurrent}. 
In contrast, the dd-CRP makes no Markovian assumptions, and efficient inference is possible through relatively straightforward Gibbs sampling in a fixed window.

%% file: conclusion.tex
\section{Conclusion}
We present a simple non-parametric model for clustering short documents (such as tweets) into storylines, which are conceptually coherent and temporally focused. 
Future work may consider learning more flexible temporal distance functions, which could potentially represent temporal periodicity or parametric models of content popularity.

%% file: ack.tex
\paragraph{Acknowledgments} We thank the reviewers for their helpful feedback. This research was supported by an award from the National Institutes for Health (R01GM112697-01), and by Google, through a Focused Research Award for Computational Journalism.